\documentclass[10pt,twocolumn,letterpaper]{article}

\usepackage{cvpr}
\usepackage{times}
\usepackage{epsfig}
\usepackage{graphicx}
\usepackage{amsmath}
\usepackage{amssymb}
\usepackage{booktabs}
\usepackage{subcaption}

\usepackage[pagebackref=true,breaklinks=true,letterpaper=true,colorlinks,bookmarks=false]{hyperref}

\cvprfinalcopy 

\def\cvprPaperID{5137} 

\ifcvprfinal\fi
\begin{document}

\title{AET vs. AED: Unsupervised Representation Learning by Auto-Encoding Transformations rather than Data}

\author{Liheng Zhang~$^\flat$\thanks{The work was done while L. Zhang was interning at Huawei Cloud.}~,~ {\bf Guo-Jun Qi}~$^\flat$$^\natural$\thanks{Corresponding author: G.-J. Qi, email: guojunq@gmail.com.}~,~Liqiang Wang$^\sharp$~,~Jiebo Luo$^\S$
\vspace{2mm}\\
$^\flat$Laboratory for MAchine Perception and LEarning (MAPLE)\\
\url{http://maple-lab.net/}\vspace{1.5mm}\\
$^\natural$Huawei Cloud,~$^\sharp$University of Central Florida,~$^\S$University of Rochester\\
{\tt\small guojunq@gmail.com}\\
{\small\url{https://github.com/maple-research-lab/AET}}
}

\maketitle

\begin{abstract}
The success of deep neural networks often relies on a large amount of labeled examples, which can be difficult to obtain in many real scenarios. To address this challenge, unsupervised methods are strongly preferred for training neural networks without using any labeled data. In this paper, we present a novel paradigm of unsupervised representation learning by Auto-Encoding Transformation (AET) in contrast to the conventional Auto-Encoding Data (AED) approach. Given a randomly sampled transformation, AET seeks to predict it merely from the encoded features as accurately as possible at the output end. The idea is the following: as long as the unsupervised features successfully encode the essential information about the visual structures of original and transformed images, the transformation can be well predicted. We will show that this AET paradigm allows us to instantiate a large variety of transformations, from  parameterized, to non-parameterized and GAN-induced ones. Our experiments show that AET greatly improves over existing unsupervised approaches, setting new state-of-the-art performances being greatly closer to the upper bounds by their fully supervised counterparts on CIFAR-10, ImageNet and Places datasets.
\end{abstract}

\section{Introduction}

\begin{figure}[t]
    \centering
    \begin{subfigure}[c]{0.45\textwidth}
        \includegraphics[width=\textwidth]{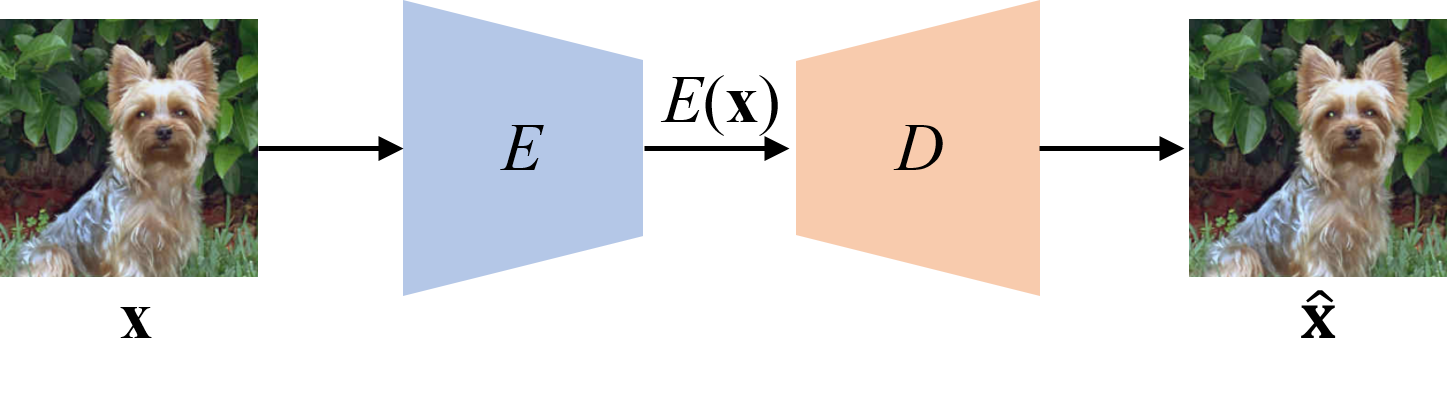}
        \caption{Auto-Encoding Data (AED)}
    \end{subfigure}\\
    ~ 
    \begin{subfigure}[c]{0.45\textwidth}
        \includegraphics[width=\textwidth]{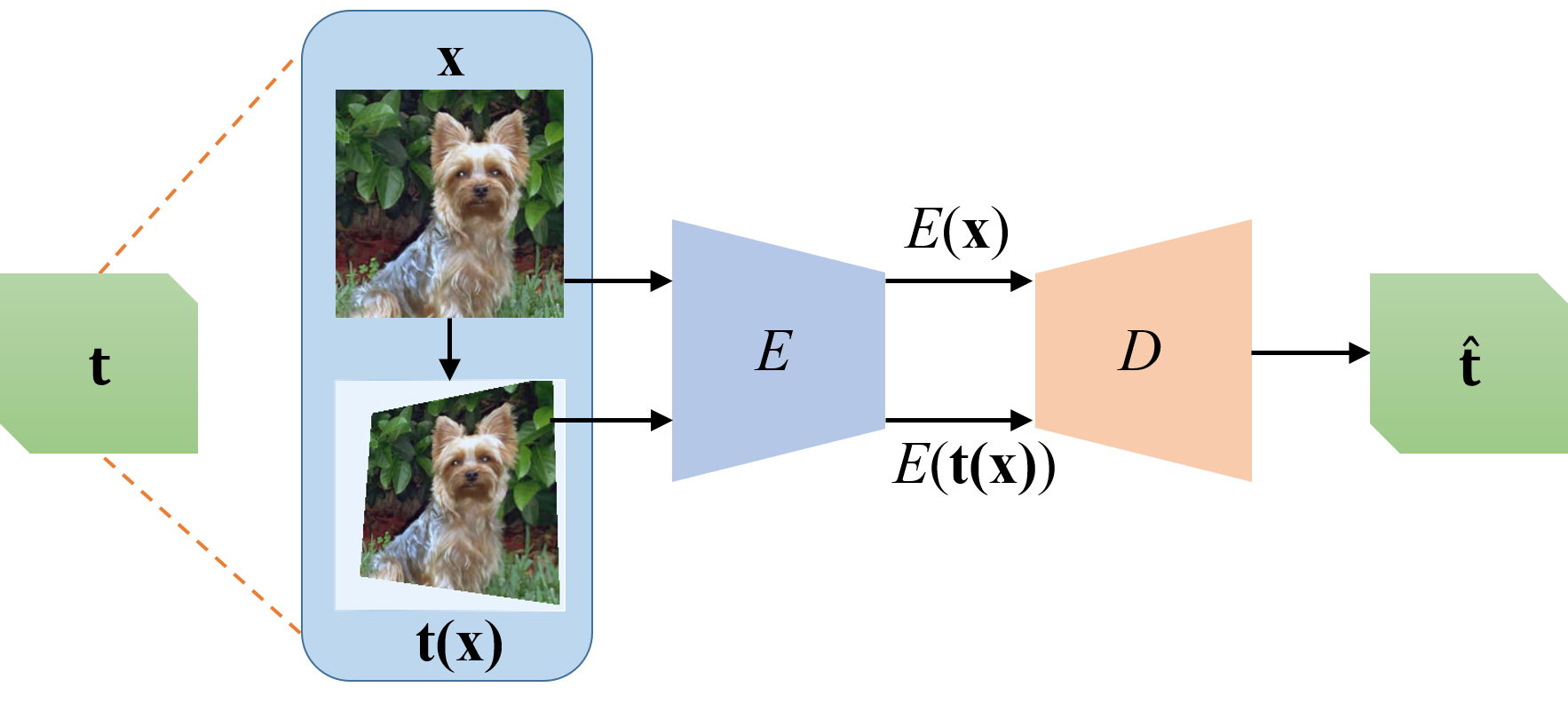}
        \caption{Auto-Encoding Transformation (AET)}
    \end{subfigure}
    \caption{An illustrative comparison between AED and AET, where AET attempts to estimate the input transformation rather than the data at the output end. This forces the encoder network $E$ to extract the features that contain the sufficient information about visual structures to decode the input transformation. }\label{fig:comparison}
\end{figure}

The great success in applying deep neural networks to image classification, object detection and semantic segmentation has inspired us to explore their full ability in a wide variety of computer vision tasks. Unfortunately, training deep neural networks often requires a large amount of labeled data to learn adequate feature representations for visual understanding tasks. This greatly limits the applicability of deep neural networks when only a limited amount of labeled data is available for training the networks.  Therefore, there has been an increasing interest in literature to learn deep feature representations in an unsupervised fashion to solve emerging visual understanding tasks with insufficient labeled data.

Among the efforts on unsupervised learning methods, the most representative ones are Auto-Encoders and Generative Adversarial Nets (GANs)~\cite{goodfellow2014generative}.  The former trains an encoder network to output feature representations with sufficient information to reconstruct input images by a paired decoder. Many variants of auto-encoders \cite{hinton2011transforming,kingma2013auto} have been proposed in literature but all of them stick to essentially the same idea of reconstructing input data at the output end, and thus we classify them into the {\em Auto-Encoding Data} (AED) paradigm illustrated in Figure~\ref{fig:comparison}(a).

On the other hand, GANs learn the feature representation in an unsupervised fashion by generating images from input noises with a pair of adversarially trained generator and discriminator. The input noises into the generator can be viewed as the feature representations of its output, since they contain necessary information to produce the corresponding images through the generator. To obtain the ``noise" feature representation for each image, an encoder can be trained to form an auto-encoder architecture with the generator as the decoder. In this way, given an input image, the encoder can directly output its noise representation producing the original image through the generator \cite{donahue2016adversarial,dumoulin2016adversarially}. This combines the strength of both AED and GAN models. Recently, these models become a popular alternative to auto-encoders in many unsupervised and semi-supervised tasks, as they can generate the {\em distribution} of photo-realistic images as a whole so that better feature representations can be derived from the trained generator.

Besides auto-encoders and GANs, various paradigms of self-supervised learning methods exist without using manually labeled data. These methods create self-supervised objectives to train the networks.
For example, Doersch et al.~\cite{doersch2015unsupervised} propose to train neural networks by predicting the relative positions of two randomly sampled patches. Mehdi and Favaro~\cite{noroozi2016unsupervised} report to train a convolutional neural network by solving Jigsaw puzzles. Image colorization has also been used as a self-supervised task to train convolutional networks in literature \cite{zhang2016colorful,larsson2016learning}.
Instead, Dosovitskiy et al.~\cite{dosovitskiy2014discriminative} train neural networks by discriminating among a set of surrogate classes artificially formed by applying various transformations to image patches, while Gidaris et al.~\cite{gidaris2018unsupervised} attempt to classify image rotations of four discrete angles. These approaches explore supervisory signals at various levels of visual structures to train networks without manually labeling data. Unsupervised features have also been extracted from videos by estimating the self-motion of moving objects between consecutive frames \cite{agrawal2015learning}.

In contrast, we are motivated to learn unsupervised feature representations by {\em Auto-Encoding Transformations} (AET) rather than the data themselves. Specifically, by sampling some operators to transform images, we seek to train auto-encoders that can directly reconstruct these operators from the learned feature representations between original and transformed images. We believe as long as the trained features are sufficiently informative, we can decode the transformations from the features that well encode visual structures of images. As compared with the conventional paradigm of Auto-Encoding Data (AED) in Figure~\ref{fig:comparison}, AET focuses on exploring dynamics of feature representations under different transformations, thereby revealing not only static visual structures but also how they would change by applying different transformations. Moreover, there is no restriction on the form of transformations applicable in the proposed AET framework. This allows us to flexibly explore a large variety of transformations, ranging from simple image warping to any  parametric and non-parametric transformations. We will demonstrate the AET representations outperform the other unsupervised models in experiments, greatly pushing the state-of-the-art unsupervised method much closer to the upper bound set by the fully supervised counterparts.

The remainder of the paper is organized as follows. We first review the related work in Section~\ref{sec:review}, and then formally present the proposed AET model in Section~\ref{sec:aet}. We conduct experiments in Section~\ref{sec:exp} to compare its performances with the other state-of-the-art unsupervised models. Finally, we summarize conclusions in Section~\ref{sec:concl}.

\section{Related Work}\label{sec:review}

{\noindent \bf Auto-Encoders.}
The use of auto-encoder architecture in learning representations in an unsupervised fashion has been extensively studied in literature \cite{hinton1994autoencoders,japkowicz2000nonlinear,vincent2008extracting}. These existing auto-encoders are all based on reconstructing the input {\em data} at the output end through a pair of encoder and decoder. The encoder acts as an extractor of features usually compactly representing the most essential information about input data, while a decoder is jointly trained to recover the input data upon the extracted features. The idea is that a good feature representation should contain sufficient information to reconstruct the input data. A wide spectrum of auto-encoders have been proposed following this paradigm of auto-encoding data (AED).
For example, the variational auto-encoder \cite{kingma2013auto} explicitly introduces probabilistic assumption about the distribution of features extracted from data. Denoising auto-encoder \cite{vincent2008extracting} aims to learn more robust representation by reconstructing original inputs from noise-corrupted inputs. Contrastive Auto-Encoder \cite{rifai2011contractive} penalizes abrupt changes of representations around given data, thus encouraging representation invariance to small perturbation on input data.  Zhang et al.~\cite{zhang2017split} present a cross-channel auto-encoder by reconstructing a subset of data channels from another subset with the cross-channel features being concatenated as data representation. Hinton et al.~\cite{hinton2011transforming} propose a transforming auto-encoder in the context of capsule nets, which is still trained in the AED fashion by minimizing the discrepancy between the reconstructed and target images.
 Conceptually, this differs from the proposed AET that aims to learn unsupervised features by directly minimizing the input and output transformations in an end-to-end auto-encoder architecture.

{\noindent \bf Generative Adversarial Nets.} Besides the auto-encoders, Generative Adversarial Nets (GANs)  become popular for training network representations of data in an unsupervised fashion.  Unlike the auto-encoders, GANs attempt to directly generate data from noises drawn from a random distribution. By viewing the sampled noises as the coordinates over the manifold of real data, one can use them as the features to represent data. For this purpose, one usually needs to train a data encoder to find the noise that can generate the input images through the GAN generator. This can be implemented by jointly training a pair of mutually inverse generator and encoder \cite{donahue2016adversarial,dumoulin2016adversarially}. A prominent characteristic of GANs that make them different from auto-encoders is they do not rely on one-to-one reconstruction of input data at the output end. Instead, they focus on discovering and generating the entire distribution of data over the underlying manifold. Recent progress has shown the promising generalization ability of regularized GANs in generating unseen data based on the Lipschitz assumption on the real data distribution \cite{qi2017loss,arjovsky2017wasserstein}, and this shows great potential of GANs in providing expressive representation of images \cite{donahue2016adversarial,dumoulin2016adversarially,edraki2018generalized}.

{\noindent \bf Self-Supervised Representation Learning.} In addition to auto-encoders and GANs,  other unsupervised learning methods explore various self-supervised signals to train deep neural networks. These self-supervised signals can be directly derived from data themselves without having to be manually labeled. For example, Doersch et al.~\cite{doersch2015unsupervised} use the relative positions of two randomly sampled patches from an image as self-supervised information to train the model. Mehdi and Favaro~\cite{noroozi2016unsupervised} propose to train a convolutional neural network by solving Jigsaw puzzles. Noroozi et al.~\cite{noroozi2017representation} learn counting features that satisfy equivalence relations between downsampled and tiled images, and Gidaris et al.~\cite{gidaris2018unsupervised} train neural networks by classifying image rotations in a discrete set. Dosovitskiy et al.~\cite{dosovitskiy2014discriminative} train CNNs by classifying a set of surrogate classes, each of which is formed by applying various transformations to an individual image. However, the resultant features could over-discriminate visually similar images as they always belong to different surrogate classes, and the training cost is much more expensive as every training example results in an individual surrogate class. 
Luo et al. \cite{luo2005image} introduce the concept of image-transform bootstrapping using transforms in the image space to augment training, testing, and both.
The idea has also been employed to train feature representations for videos through the self-motion of moving objects \cite{agrawal2015learning}. In summary, this type of approaches train networks using various self-supervised objectives instead of manually labeled data.

\section{AET: The Proposed Approach}\label{sec:aet}
We elaborate on the proposed paradigm of auto-encoding transformations (AET) in this section.
First, we will formally present the formulation of AET in Section~\ref{sec:form}. Then we will instantiate AET with different genres of transformations in Section~\ref{sec:ex}.

\subsection{The Formulation}\label{sec:form}
Suppose that we sample a transformation $\mathbf t$ from a distribution $\mathcal T$ (e.g., image warping, projective transformation and even GAN-induced transformation, c.f.~Section~\ref{sec:ex} for more details). It is applied to an image $\mathbf x$ drawn from a data distribution $\mathcal X$, resulting in the transformed version $\mathbf t(\mathbf x)$ of $\mathbf x$.

Our goal is to learn an encoder $E: \mathbf x \mapsto E(\mathbf x)$,
which aims to extract the representation $E(\mathbf x)$ for a sample $\mathbf x$.
Meanwhile, we wish to learn a decoder $D: \left[E(\mathbf x), E(\mathbf t(\mathbf x))\right]\mapsto \hat {\mathbf t}$, which gives an estimate $\hat{\mathbf t}$ of input transformation by decoding from the encoded representations of original and transformed images. Since the prediction on the input transformation is made through the encoded features rather than the original and transformed images, it forces the model to extract expressive features as a proxy to represent images.

The learning problem of Auto-Encoding Transformations (AET) now boils down to jointly training the feature encoder $E$ and the transformation decoder $D$. To this end, let us choose a loss function $\ell(\mathbf t, \hat{\mathbf t})$ that quantifies the difference between a transformation $\mathbf t$ and its estimate $\hat{\mathbf t}$. Then the AET can be solved by minimizing this loss as

$$
\min_{E,D} \mathop\mathbb E\limits_{\mathbf t \sim \mathcal T, \mathbf x\sim \mathcal X}\ell(\mathbf t, \hat{\mathbf t})
$$
where the transformation estimate $\hat{\mathbf t}$ is a function of the encoder $E$ and the decoder $D$ such that
$$
\hat{\mathbf t}= D\left[E(\mathbf x), E(\mathbf t(\mathbf x))\right],
$$
and the expectation $\mathbb E$ is taken over the sampled transformations and data.
Like in training other deep neural networks, the network parameters of $E$ and $D$ are jointly updated over mini-batches by back-propagating the gradient of the loss $\ell$.

\subsection{The AET Family}\label{sec:ex}
A large variety of transformations can be easily incorporated into the AET formulation. Here we discuss three genres, parameterized, GAN-induced and non-parameterized transformations, to instantiate the AET models.

{\noindent\bf Parameterized Transformations.}
Suppose that we have a family of transformations $\mathcal T=\{\mathbf t_{\boldsymbol\theta}|\boldsymbol\theta\sim \boldsymbol\Theta\}$ with their parameters $\boldsymbol \theta$ sampled from a distribution $\boldsymbol\Theta$. This equivalently defines a distribution of parameterized transformations, where each transformation can be represented by its parameter and the loss $\ell(\mathbf t_{\boldsymbol\theta}, \mathbf t_{\hat{\boldsymbol\theta}})$ between transformations can be captured by the difference in terms of their parameters.
For example, many transformations such as affine and projective transformations can be represented by a parameterized matrix $M(\boldsymbol\theta)\in \mathbb R^{3\times 3}$ between homogeneous coordinates of images before and after transformations. Such a matrix captures the change of geometric structures caused by a given transformation, and thus
it is straightforward to define $\ell(\mathbf t_{\boldsymbol\theta},\mathbf t_{\boldsymbol{\hat\theta}})=\dfrac{1}{2}\|M(\boldsymbol\theta)-M(\hat{\boldsymbol\theta})\|^2_2$ to model the difference between the target and the estimated transformations.
In the experiments, we will compare
different instances of parameterized transformations in this category and demonstrate they can yield competitive performances on training AET.

{\noindent\bf GAN-Induced Transformations.} One can choose other forms of transformations without explicit geometric implications like the affine and the projective transformations.
Let us consider a GAN generator that transforms an input over the manifold of real images. For example, in \cite{qi2018global}, a local generator $G(\mathbf x, \mathbf z)$ is learned with a sampled random noise $\mathbf z$ that parameterizes the underlying transformation around a given image $\mathbf x$. This effectively defines a GAN-induced transformation such that $\mathbf t_\mathbf z(\mathbf x) = G(\mathbf x, \mathbf z)$ with the transformation parameter $\mathbf z$. One can directly choose the loss $\ell(\mathbf t_\mathbf z, \mathbf t_{\hat{\mathbf z}})=\dfrac{1}{2}\|\mathbf z-\hat{\mathbf z}\|_2^2$ between noise parameters, and train a network $D$ to decode the parameter $\hat{\mathbf z}$ from the features $E(\mathbf x)$ and $E(\mathbf t_\mathbf z(\mathbf x))$ by the encoder network $E$. Compared with the classical transformations that change low-level appearance and geometric structures in images, the GAN-induced transformations can change high-level semantics in images. For example, the GANs have demonstrated their ability of manipulating attributes such as ages, hairs, genders and wearing glasses in facial images as well as changing the furniture layout in bedroom images \cite{radford2015unsupervised}. This enables AET to explore a richer family of transformations to learn more expressive representations.

{\noindent \bf Non-Parametric Transformations.} Even if a transformation $\mathbf t\in\mathcal T$ is hard to parameterize, we can still define the loss $\ell(\mathbf t, \hat{\mathbf t})$ by measuring the average difference between the transformations of randomly sampled images. Formally,
\begin{equation}\label{eq:nloss}
\ell(\mathbf t, \hat{\mathbf t})=\mathop\mathbb E\limits_{\mathbf x\sim \mathcal X} {\rm dist}(\mathbf t(\mathbf x), \hat{\mathbf t}(\mathbf x))
\end{equation}
where ${\rm dist}(\cdot,\cdot)$ is a distance between two transformed images, and the expectation is taken over random samples.  For an input non-parametric transformation $\mathbf t$, we also need a decoder network that outputs a transformation $\hat{\mathbf t}$ to estimate the input transformation. This can be done by choosing a parameterized transformation $\mathbf t_{\hat{\boldsymbol\theta}}$ as $\hat{\mathbf t}$ to estimate $\mathbf t$. Although the non-parametric $\mathbf t$ may not fall in the space of parameterized transformations, such an approximation should be enough for unsupervised learning since our ultimate goal is not to obtain an accurate estimate of input transformation; instead, we aim at learning a good feature representation to give us the best estimate that can be achieved in the parameterized transformation space.

Note that parameterized transformations can also be plugged into Eq.~(\ref{eq:nloss}) to train the corresponding AET by minimizing this loss function. However, in experiments, we find the performance is not as good as the AET trained with the parameter-based loss. This is probably caused by the fact that the loss (\ref{eq:nloss}) cannot accurately reflect the actual difference between transformations unless a sufficiently large number of images are sampled. Thus, we suggest using the parameter-based loss for the AET with parameterized transformations.


We have shown that a wide spectrum of transformations can be adopted in training AET.
In this paper, we will focus on the parameterized transformations as they do not involve training extra models like GAN-induced transformations, or require choosing auxiliary transformations to approximate non-parametric forms. This allows us to make a straightforward and fair comparison with the unsupervised methods in literature as shown in the experiments. Moreover, the GAN-induced transformations greatly rely on the quality of transformed images, but existing GAN models are still unable to generate high-quality images with fine-grained details at a high resolution. Thus, we leave it in future to study the GAN-induced and non-parametric transformations for training the AET representations.

\vspace{-2mm}
\section{Experiments}\label{sec:exp}
\vspace{-2mm}
In this section, we evaluate the proposed AET model on the CIFAR-10, ImageNet and Places datasets by comparing it against different unsupervised methods.
Unsupervised learning is usually evaluated indirectly based on the classification performance by using the learned representations. For the sake of fair comparison, we follow the test protocols widely adopted in literature.

\vspace{-2mm}
\subsection{CIFAR-10 Experiments}
\vspace{-2mm}
First, we evaluate the AET model on the CIFAR-10 dataset. We consider two different transformations -- affine and projective transformations -- to train AET, and name the resultant models AET-affine and AET-project for brevity, respectively.

\begin{figure}[t]
\begin{center}
   \includegraphics[width=1.0\linewidth]{{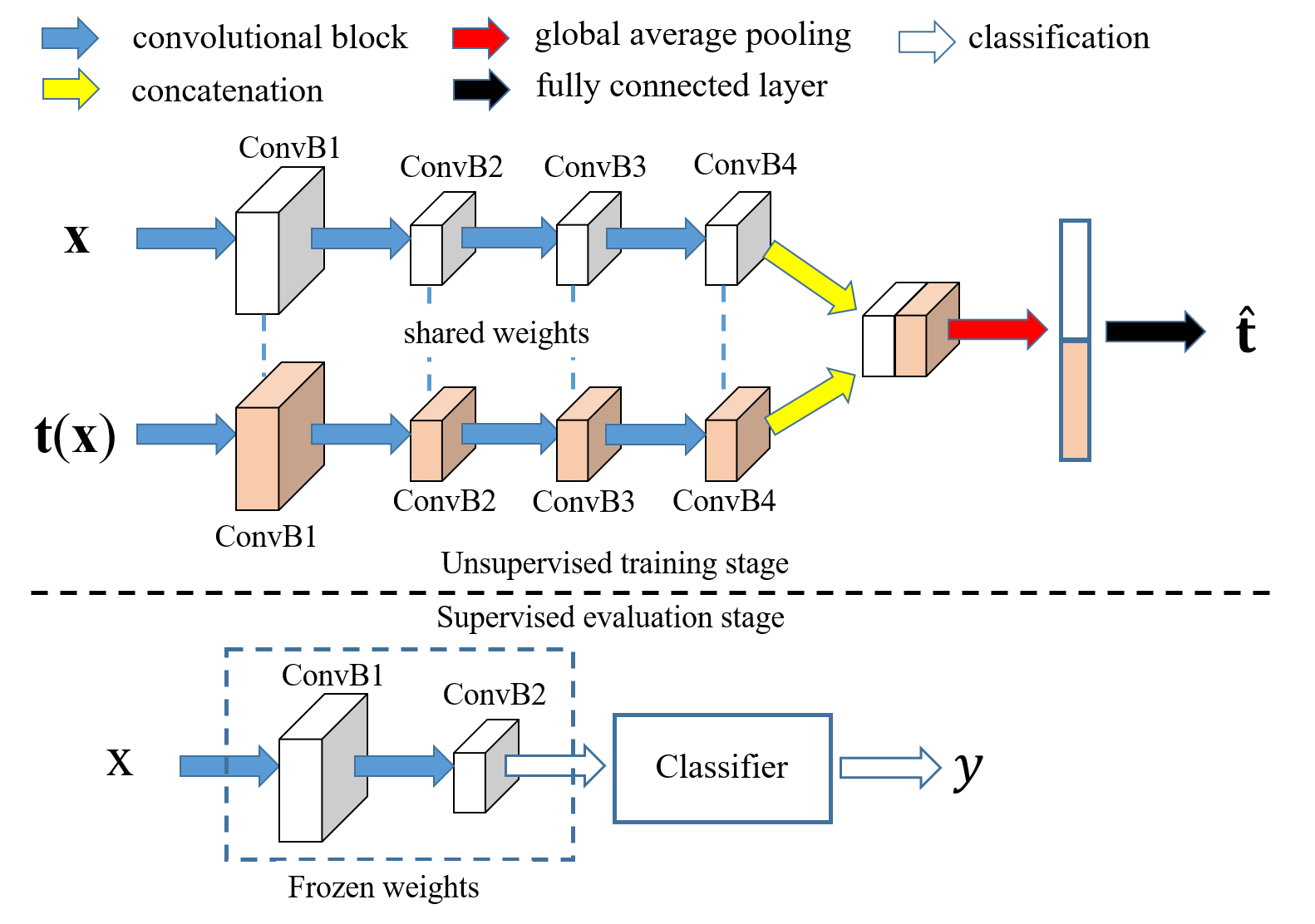}}
\end{center}
   \caption{An illustration of the network architectures for training and evaluating AET on the CIFAR-10 dataset.}
\label{fig:cifar_arch}
\vspace{-5mm}
\end{figure}

\vspace{-2mm}
\subsubsection{Architecture and Implementation Details}
\vspace{-2mm}
To make a fair and direct comparison with existing unsupervised models, we adopt the Network-In-Network (NIN) architecture that has shown competitive performance previously on the CIFAR-10 dataset for the unsupervised learning task \cite{gidaris2018unsupervised}.  As illustrated in the top of Figure~\ref{fig:cifar_arch}, the NIN consists of four convolutional blocks, each of which contains three convolutional layers.
AET has two NIN branches, each taking the original and the transformed images as its input, respectively. The output features of the forth block of two branches are concatenated and average-pooled to form a $384$-d feature vector. Then an output layer follows to predict the parameters of input transformation. The two branches share the same network weights, and are used as the encoder network producing the feature representations for input images.

The AET networks are trained by SGD with a batch size of $512$ original images and their transformed counterparts. Momentum and weight decay are set to $0.9$ and $5\times 10^{-4}$. The learning rate is initialized to $0.1$ and scheduled to drop by a factor of $5$ after $240$, $480$, $640$, $800$ and $1,000$ epochs. The model is trained for $1,500$ epochs in total. For AET-affine, the affine transformation is a composition of a random rotation with $[-180^\circ, 180^\circ]$, a random translation by $\pm 0.2$ of image height and width in both vertical and horizontal directions, and a random scaling factor of $[0.7, 1.3]$, along with a random shearing of $[-30^\circ,30^\circ]$ degree.  For the AET-projective, the projective transformation is formed by
randomly translating four corners of an image in both horizontal and vertical directions by $\pm 0.125$ of its height and width,  after it is randomly scaled by $[0.8, 1.2]$ and rotated by $0^\circ, 90^\circ, 180^\circ,$ or $270^\circ$. We compare the results for both models below, and demonstrate both outperform the other existing models and AET-project performs better than AET-affine.

\subsubsection{Evaluation Protocol}

To evaluate the quality of the representation by an unsupervised model, a classifier is usually trained upon the learned features. Specifically, in our experiments on CIFAR-10, we follow the existing evaluation protocols \cite{oyallon2015deep,dosovitskiy2014discriminative,radford2015unsupervised,oyallon2017scaling,gidaris2018unsupervised} by building a classifier on top of the second convolutional block. See the bottom of Figure~\ref{fig:cifar_arch}, where the first two blocks are frozen while the classifier on top of them is trained with labeled examples.

We evaluate the classification results by using the AET features with both model-based and model-free classifiers.  For the model-based classifier, we follow \cite{gidaris2018unsupervised} by training a non-linear classifier with three Fully-Connected (FC) layers -- each of the two hidden layers has $200$ neurons with batch-normalization and ReLU activations, and the output layer is a soft-max layer with ten neurons each for an image class. Alternatively, we also test a convolutional classifier upon the unsupervised features by adding a third NIN block whose output feature map is averaged pooled and connected to a linear soft-max classifier.

Moreover, we also test the model-free KNN classifier based on the averaged-pooled output features from the second convolutional block. The KNN classifier has an advantage without need to train a model with labeled examples.  This makes a more direct evaluation on the quality of unsupervised feature representation at the evaluation stage.

\subsubsection{Results}

\begin{table}
\caption{Comparison between unsupervised feature learning methods on CIFAR-10. The fully supervised NIN and the random Init. + conv have the same three-block NIN architecture, but the first is fully supervised while the second is trained on top of the first two blocks that are randomly initialized and stay frozen during training.}\label{tab01}
\centering
 \begin{tabular}{l|c} \toprule
Method&Error rate\\ \midrule
Supervised NIN (Lower Bound)&7.20  \\
Random Init. + conv (Upper Bound)&27.50  \\ \midrule
Roto-Scat + SVM \cite{oyallon2015deep} &17.7 \\
ExamplarCNN \cite{dosovitskiy2014discriminative} &15.7 \\
DCGAN \cite{radford2015unsupervised}&17.2 \\
Scattering \cite{oyallon2017scaling}&15.3\\
RotNet + FC \cite{gidaris2018unsupervised}&10.94\\
RotNet + conv \cite{gidaris2018unsupervised}&8.84\\ \midrule
(Ours) AET-affine + FC &9.77\\
(Ours) AET-affine + conv &8.05\\
(Ours) AET-project + FC &\textbf{9.41}\\
(Ours) AET-project + conv &\textbf{7.82}\\ \bottomrule
\end{tabular}
\end{table}

\begin{table*}
\caption{Comparison of RotNet vs. AETs on CIFAR-10 with different classifiers on top of learned representations for evaluation. The RotNet is chosen as the baseline since it has the exactly same architecture for the unsupervised training. Here $n$-FC denotes a $n$-layer fully connected (FC) classifier, and the KNN is obtained with $K=10$ nearest neighbors. The numbers in parentheses are the {\it relative} reduction in error rates w.r.t. the RotNet baseline.}\label{tab02}
\centering
 \begin{tabular}{c|ccccc} \toprule
   &KNN&1-FC&2-FC&3-FC&conv\\ \midrule
RotNet baseline~\cite{gidaris2018unsupervised}&24.97 &18.21&11.34&10.94 &8.84 \\
AET-affine&23.07 ($\downarrow$7.6\%) &17.16 ($\downarrow$ 5.8\%)&9.77 ($\downarrow$ 13.8\%)&10.16 ($\downarrow$ 7.1\%)&8.05($\downarrow$ 8.9\%)\\ \
AET-project &\textbf{22.39} ($\downarrow$ 10.3\%)&\textbf{16.65} ($\downarrow$ 8.6\%)&\textbf{9.41} ($\downarrow$ 17.0\%)&\textbf{9.92} ($\downarrow$ 9.3\%)&\textbf{7.82}($\downarrow$ 11.5\%) \\ \bottomrule
\end{tabular}
\end{table*}


\begin{figure}[t]
\begin{center}
   \includegraphics[width=0.66\linewidth]{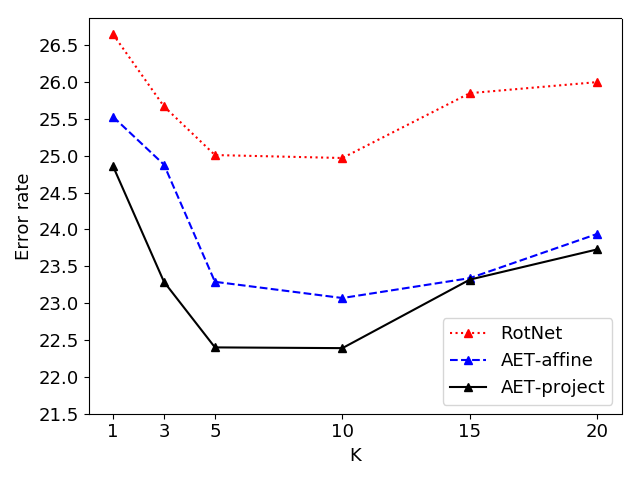}
\end{center}
   \caption{The comparison of the KNN error rates by different models with varying numbers $K$ of nearest neighbors on CIFAR-10.}
\label{fig:knn_curve}
\vspace{-5mm}
\end{figure}

In Table~\ref{tab01}, we compare the AET models with both fully supervised and unsupervised methods on CIFAR-10. First, we note that the unsupervised AET-project with the convolutional classifier almost achieves the same error rate as its fully supervised NIN counterpart with four convolutional blocks ($7.82\%$ vs. $7.2\%$). This is a remarkable result demonstrating AET is capable of training unsupervised features with a much narrower gap of performance to its supervised counterpart on CIFAR-10.

Moreover, the AETs outperform the other unsupervised methods in Table~\ref{tab01}. For example, ExamplarCNN also applies various transformations to images, including rotations, translations, scaling and even more such as manipulating contrasts and colors. Then it trains unsupervised CNNs by classifying the resultant surrogate classes each containing all transformed versions of an individual images. Compared with ExamplarCNN \cite{dosovitskiy2014discriminative}, AET still has a significant lead in error rate, implying it can explore the image transformations more effectively in training unsupervised networks.

It is worth pointing out on CIFAR-10, the other reported methods \cite{oyallon2015deep,dosovitskiy2014discriminative,radford2015unsupervised,oyallon2017scaling,gidaris2018unsupervised} are usually based on different unsupervised networks and supervised classifiers for evaluation,  making it difficult to make a direct comparison between them. The results still suggest that the state-of-the-art performances can be reached by AETs, as their error rates are very close to the pre-assumptive lower bound set by the fully supervised counterpart.

Indeed, one can choose the RotNet in Table~\ref{tab01} as the baseline for comparison as it is trained with the same network and classifier as the AETs. Thus we can make a fair comparison directly. From the results,  AETs successfully beat the RotNet with both fully connected (FC) and convolutional classifiers on top of the learned representations.
We also compare AETs with this baseline when they are trained with the KNN classifier and varying FC layers in Table~\ref{tab02}. The results show that AET-project can consistently achieve the smallest errors no matter which classifiers are used. In Figure~\ref{fig:knn_curve}, we also compare the KNN results with varying number of nearest neighbors. Again, AET-project performs the best without involving any labeled examples. The model-free KNN results suggest the AET model has an advantage when no labels are available in training classifiers upon the unsupervised features.

For the following ImageNet experiments, many existing methods have been compared in literature with the same unsupervised AlexNet architecture as well as the classifiers upon it for the evaluation. We will make a fair comparison directly, and the results show that AET still greatly outperforms the other unsupervised methods.

\subsection{ImageNet Experiments}

\begin{table}
\caption{Top-1 accuracy with non-linear layers on ImageNet. AlexNet is used as backbone to train the unsupervised models. After unsupervised features are learned, nonlinear classifiers are trained on top of Conv4 and Conv5 layers with labeled examples to compare their performances. We also compare with the fully supervised models and random models that give upper and lower bounded performances. For a fair comparison, only a single crop is applied in AET and no dropout or local response normalization is applied during the testing. }\label{tab04}
\centering
 \begin{tabular}{l|cc} \toprule
Method&Conv4 &Conv5\\ \midrule
ImageNet Labels \cite{bojanowski2017unsupervised}(Upper Bound)&59.7&59.7  \\
Random \cite{noroozi2017representation} (Lower Bound)&27.1 &12.0  \\ \midrule
Tracking \cite{wang2015unsupervised} &38.8&29.8 \\
Context \cite{doersch2015unsupervised} &45.6&30.4 \\
Colorization \cite{zhang2016colorful}&40.7&35.2 \\
Jigsaw Puzzles \cite{noroozi2016unsupervised}&45.3&34.6\\
BiGAN \cite{donahue2016adversarial}&41.9&32.2\\
NAT \cite{bojanowski2017unsupervised}&-&36.0\\
DeepCluster \cite{caron2018deep} &-&44.0\\
RotNet \cite{gidaris2018unsupervised}&50.0&43.8\\\midrule
(Ours) AET-project &\textbf{53.2}&\textbf{47.0}\\\bottomrule
\end{tabular}
\end{table}

\begin{table*}
\caption{Top-1 accuracy with linear layers on ImageNet. AlexNet is used as backbone to train the unsupervised models under comparison. A $1,000$-way linear classifier is trained upon various convolutional layers of feature maps that are spatially resized to have about $9,000$ elements. Fully supervised and random models are also reported to show the upper and the lower bounds of unsupervised model performances. Only a single crop is used and no dropout or local response normalization is used during testing for the AET, except the models denoted with * where ten crops are applied to compare results.}\label{tab05}
\centering
 \begin{tabular}{l|ccccc} \toprule
Method&Conv1 &Conv2&Conv3&Conv4&Conv5\\ \midrule
ImageNet Labels (Upper Bound) \cite{gidaris2018unsupervised}&19.3&36.3&44.2&48.3&50.5  \\
Random (Lower Bound)\cite{gidaris2018unsupervised} &11.6 &17.1&16.9&16.3&14.1  \\
Random rescaled \cite{krahenbuhl2015data}(Lower Bound)&17.5 &23.0&24.5&23.2&20.6  \\\midrule
Context \cite{doersch2015unsupervised} &16.2&23.3&30.2&31.7&29.6 \\
Context Encoders \cite{pathak2016context}&14.1&20.7&21.0&19.8&15.5 \\
Colorization\cite{zhang2016colorful}&12.5&24.5&30.4&31.5&30.3\\
Jigsaw Puzzles \cite{noroozi2016unsupervised}&18.2&28.8&34.0&33.9&27.1\\
BiGAN \cite{donahue2016adversarial}&17.7&24.5&31.0&29.9&28.0\\
Split-Brain \cite{zhang2017split}&17.7&29.3&35.4&35.2&32.8\\
Counting \cite{noroozi2017representation}&18.0&30.6&34.3&32.5&25.7\\
RotNet \cite{gidaris2018unsupervised}&18.8&31.7&38.7&38.2&36.5\\\midrule
(Ours) AET-project &\textbf{19.2}&\textbf{32.8}&\textbf{40.6}&\textbf{39.7}&\textbf{37.7}\\\midrule\midrule
DeepCluster* \cite{caron2018deep} &13.4&32.3&41.0&39.6&38.2\\\midrule
(Ours) AET-project* &\textbf{19.3}&\textbf{35.4}&\textbf{44.0}&\textbf{43.6}&\textbf{42.4}\\
\bottomrule
\end{tabular}\\
\end{table*}

We further evaluate the performance by AET on the ImageNet dataset. The AlexNet is used as the backbone to learn the unsupervised features. As shown by the results on CIFAR-10, the projective transformation has better performance on training the AET model, and thus we report the AET-project results here.

{\noindent \bf Architectures and Training Details.}
Two AlexNet branches with shared parameters are created with original and transformed images as inputs respectively to train unsupervised AET-project. The $4,096$-d output features from the second last fully connected layer in two branches are concatenated and fed into the output layer producing eight projective transformation parameters. We still use SGD to train the network, with a batch size of $768$ images and their corresponding transformed version, a momentum of $0.9$, a weight decay of $5\times 10^{-4}$. The initial learning rate is set to $0.01$, and it is dropped by a factor of $10$ at epoch 100 and 150. AET is trained for $200$ epochs in total. Finally, the projective transformations applied are randomly sampled in the same fashion as on CIFAR-10.


{\noindent\bf Results.} First we report the Top-1 accuracies of compared methods in Table~\ref{tab04} on ImageNet by following the evaluation protocol in \cite{noroozi2016unsupervised}. Two settings are adopted for evaluation -- Conv4 and Conv5 denote to train the remaining part of AlexNet on top of Conv4 and Conv5 with the labeled data, while all the bottom convolutional layers up to Conv4 and Conv5 are frozen after they are trained in an unsupervised fashion. For example, in the Conv4 setting, Conv5 and three fully connected layers are trained on the labeled examples, including the last $1000$-way output layer. From the results, in both settings, the AET model successfully beats the other compared unsupervised models.
In particular, among the compared models is the BiGAN ~\cite{donahue2016adversarial} that trains a GAN-based unsupervised model, and learns a data-based auto-encoder as well to map an image to an unsupervised representation. Thus, it can be seen as combing the strengths of both GAN and AED models. The results show AET outperforms BiGAN by a significant lead, suggesting its advantage over the GAN and AED paradigms at least in this experiment setting.

We also compare with the fully supervised models that give the upper bounded performance by training the entire AlexNet with all labeled data. The classifiers of random models are trained on top of Conv4 and Conv5 with randomly sampled weights, and they set up the lower bounded performance. From the comparison, the AET models greatly narrow the performance gap to the upper bound -- the gap to the upper bound Top-1 accuracy has been decreased from $9.7\%$ and $15.7\%$ by RotNet and DeepCluster on Conv4 and Conv5, respectively, to $6.5\%$ and $12.7\%$ by AET, which is {\it relatively} narrowed by $33\%$ and $19\%$,  respectively.

Moreover, we also follow the testing protocol adopted in \cite{zhang2017split} to compare the models by training a $1,000$-way linear classifier on top of different numbers of convolutional layers in Table~\ref{tab05}.  Again, AET obtains the best accuracy among all the compared unsupervised models.

\subsection{Places Experiments}

\begin{table*}
\caption{Top-1 accuracy on the Places dataset with linear layers. A $205$-way logistic regression classifier is trained on top of various layers of feature maps that are spatially resized to have about $9,000$ elements. All unsupervised features are pre-trained on the ImageNet dataset, which are frozen when training the logistic regression layer with Places labels. We also compare them with fully-supervised networks trained with Places Labels and ImageNet labels, along with random models. The highest accuracy values are in bold and the second highest accuracy values are underlined. }\label{tab06}
\centering
 \begin{tabular}{l|ccccc} \toprule
Method&Conv1 &Conv2&Conv3&Conv4&Conv5\\ \midrule
Places labels \cite{zhou2014learning}&22.1&35.1&40.2&43.3&44.6 \\
ImageNet labels&22.7&34.8&38.4&39.4&38.7\\
Random &15.7 &20.3&19.8&19.1&17.5  \\
Random rescaled \cite{krahenbuhl2015data}&21.4 &26.2&27.1&26.1&24.0  \\
\midrule
Context \cite{doersch2015unsupervised} &19.7&26.7&31.9&32.7&30.9 \\
Context Encoders \cite{pathak2016context}&18.2&23.2&23.4&21.9&18.4 \\
Colorization\cite{zhang2016colorful}&16.0&25.7&29.6&30.3&29.7\\
Jigsaw Puzzles \cite{noroozi2016unsupervised}&\underline{23.0}&31.9&35.0&34.2&29.3\\
BiGAN \cite{donahue2016adversarial}&22.0&28.7&31.8&31.3&29.7\\
Split-Brain \cite{zhang2017split}&21.3&30.7&34.0&34.1&32.5\\
Counting \cite{zhang2017split}&\textbf{23.3}&\textbf{33.9}&\underline{36.3}&\underline{34.7}&29.6\\
RotNet \cite{gidaris2018unsupervised}&21.5&31.0&35.1&34.6&\underline{33.7}\\
 \midrule
(Ours) AET-project &22.1&\underline{32.9}&\textbf{37.1}&\textbf{36.2}&\textbf{34.7}\\\bottomrule
\end{tabular}
\end{table*}

We also conduct experiments on the Places dataset. As shown in Table~\ref{tab06}, we evaluate unsupervised models that are pretrained on the ImageNet dataset. Then a single-layer logistic regression classifier is trained on top of different layers of feature maps with Places labels. Thus, we assess the generalizability of unsupervised features from one dataset to another. Our models are still based on AlexNet variants like those used in the ImageNet experiments.
We also compare with the fully supervised models trained with the Places labels and ImageNet labels,as well as the random networks. The results show the AET models outperform the other unsupervised models in most of cases, except on Conv1 and Conv2, Counting \cite{zhang2017split} performs slightly better.

\subsection{Analysis of Predicated Transformations}
\begin{figure}[t]
    \centering
    \begin{subfigure}[c]{0.24\textwidth}
        \includegraphics[width=\textwidth]{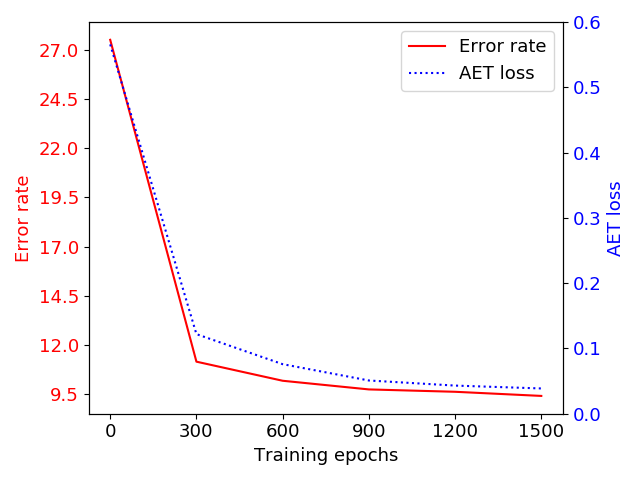}
        \caption{CIFAR-10}
    \end{subfigure}
    ~ 
    \begin{subfigure}[c]{0.24\textwidth}
        \includegraphics[width=\textwidth]{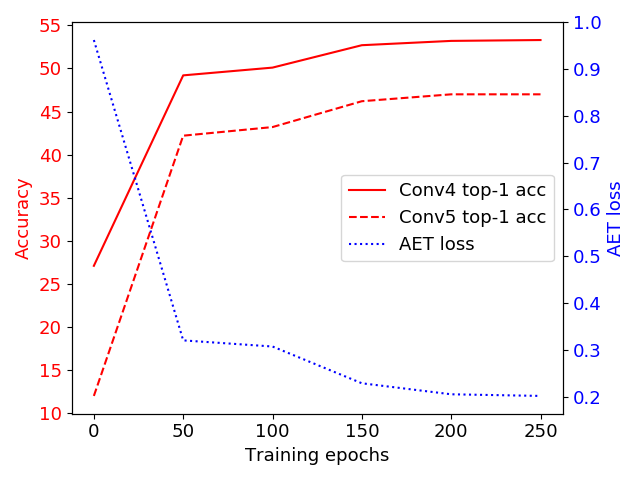}
        \caption{ImageNet}
    \end{subfigure}
    \caption{Error rate(top-1 accuracy) vs. AET loss over epochs on the CIFAR-10 and ImageNet datasets. }\label{fig:curves}
    \vspace{-2mm}
\end{figure}

Although our ultimate goal is to learn good representations of images, it is insightful to look into the accuracy of predicting transformations and its relation with the supervised classification performance. As illustrated in Figure~\ref{fig:curves}, the trend of transformation prediction loss (i.e. the AET loss being minimized to train the model) is well aligned with that of classification error and Top-1 accuracy on CIFAR-10 and ImageNet. This suggests that better prediction of transformations is a good surrogate of better classification result by using the learned features. This justifies our choice of AET to supervise the learning of feature representations.

\begin{figure}[t]
\begin{center}
   \includegraphics[width=1.0\linewidth]{{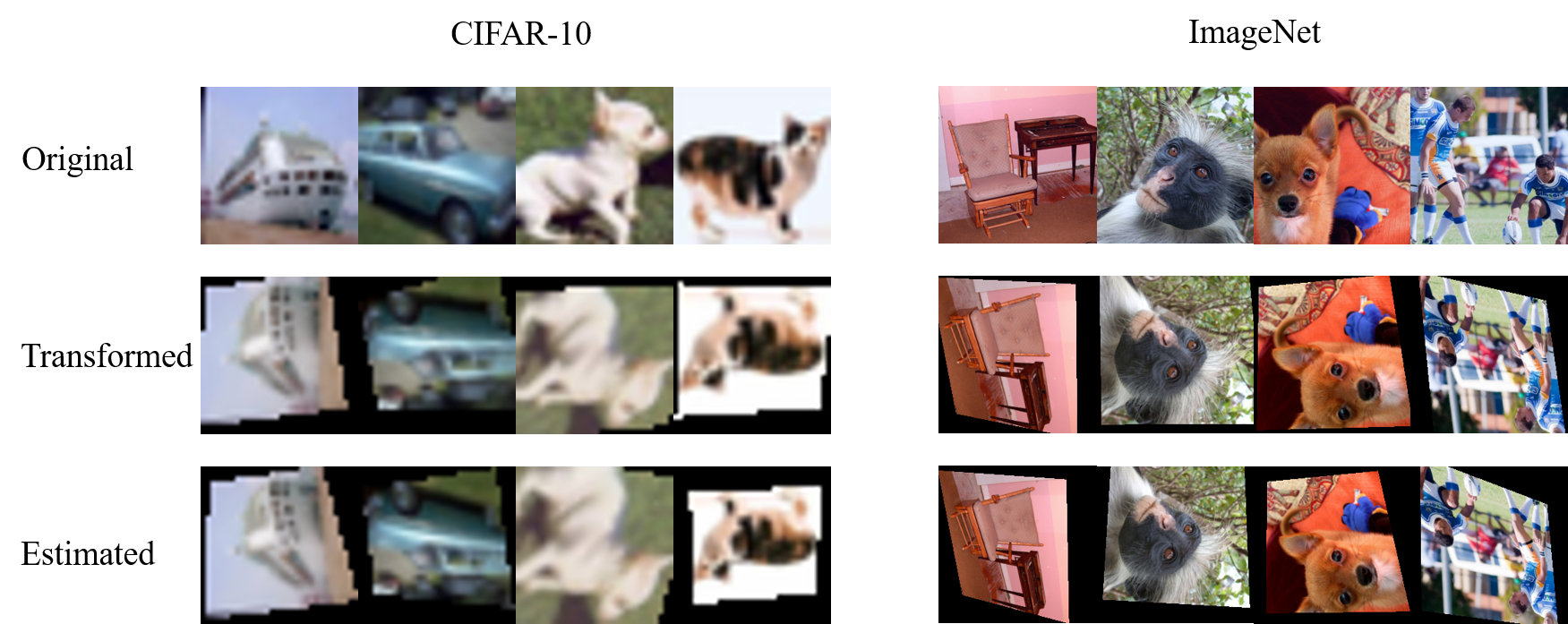}}
\end{center}
   \caption{Some examples of original images (top), along with the counterparts of input (middle) and predicted (bottom) transformations by the AET model.}
\label{fig:pred_transform}
\vspace{-2mm}
\end{figure}

In Figure~\ref{fig:pred_transform}, we also compare some examples of original images, along with the transformed images at the input and the output ends of the AET model. These examples show how well the model can decode the transformations from the encoded image features, thereby delivering unsupervised representations that offer competitive performances on classifying images in our experiments.

\section{Conclusions}\label{sec:concl}
In this paper, we present a novel Auto-Encoding Transformation (AET) paradigm for unsupervised training of neural networks in contrast to the conventional Auto-Encoding Data (AED) approach. By estimating randomly sampled transformations at output end, AET forces the encoder to learn good representations so that they contain sufficient information about visual structures of both the original and transformed images. We demonstrate that a wide variety of transformations can be easily incorporated into this framework and the experiment results demonstrate substantial improvements over the state-of-the-art performances,  significantly narrowing the gap with the fully supervised counterparts in literature.

\section{Acknowledgement}
This work was done during Liheng Zhang was interning at Huawei Cloud, Seattle WA, while the idea was conceived and formulated by Guo-Jun Qi.




{\small
\bibliographystyle{ieee}
\bibliography{egbib}
}

\end{document}